\documentclass[10pt,twocolumn,letterpaper]{article}

\usepackage{iccv}
\usepackage{times}
\usepackage{epsfig}
\usepackage{graphicx}
\usepackage{amsmath}
\usepackage{amssymb}
\usepackage{tabularx}

\newcolumntype{Y}{>{\centering\arraybackslash}X}

\usepackage[pagebackref=true,breaklinks=true,letterpaper=true,colorlinks,bookmarks=false]{hyperref}

\iccvfinalcopy 

\def\iccvPaperID{****} 

\ificcvfinal\fi

\begin{document}

\title{Feature-compatible Progressive Learning for Video Copy Detection}

\author{
Wenhao Wang\\
ReLER, University of Technology Sydney\\
{\tt\small wangwenhao0716@gmail.com}
\and
Yifan Sun\\
Baidu Inc.\\
{\tt\small sunyifan01@baidu.com}
\and
Yi Yang\\
Zhejiang University\\
{\tt\small yangyics@zju.edu.cn}
}

\maketitle
\ificcvfinal\thispagestyle{empty}\fi

\begin{abstract}
Video Copy Detection (VCD) has been developed to identify instances of unauthorized or duplicated video content. This paper presents our second place solutions to the Meta AI Video Similarity Challenge (VSC22), CVPR 2023. In order to compete in this challenge, we propose Feature-Compatible Progressive Learning (FCPL) for VCD. FCPL trains various models that produce mutually-compatible features, meaning that the features derived from multiple distinct models can be directly compared with one another. We find this mutual compatibility enables feature ensemble. By implementing progressive learning and utilizing labeled ground truth pairs, we effectively gradually enhance performance. Experimental results demonstrate the superiority of the proposed FCPL over other competitors. Our code is available at \href{https://github.com/WangWenhao0716/VSC-DescriptorTrack-Submission}{VSC-Descriptor} and \href{https://github.com/WangWenhao0716/VSC-MatchingTrack-Submission}{VSC-Matching}.
\end{abstract} 

\section{Introduction}
Video Copy Detection (VCD) refers to the process of identifying duplicate or near-duplicate videos within an extensive collection of videos. The primary objective of VCD is to detect instances of video piracy or unauthorized usage of copyrighted materials. During CVPR 2023, Meta AI organized a competition called the Video Similarity Challenge (VSC22), which featured both a descriptor and a matching track. In the descriptor track, participants were required to generate useful 512-dimensional vector representations of videos. Meanwhile, in the matching track, competitors aimed to develop a model that directly identifies specific clips in a query video and matches them to corresponding clips within one or more videos in a large reference video corpus. \par

This report summarizes our proposed method, Feature-Compatible Progressive Learning (FCPL), which is applicable to both tracks. Our FCPL approach draws inspiration from ISC21-winning solutions (FOSSL \cite{Dong2021FOSSL} and CNNCL \cite{yokoo2021contrastive}) and primarily builds upon our previous work (BoT \cite{wang2021bag}, D$^2$LV \cite{wang2021d}, and ASL \cite{wang2023benchmark}). In feature-compatible learning, we initially train a base network and utilize the trained network to obtain the feature distribution of original images (those without transformations). During the training of new networks, our objective is to align the features extracted by the new networks with the feature distribution obtained from the base network. By maintaining a fixed feature distribution for the original images, we can achieve ensemble by averaging the features acquired by different networks. The training of the base and new networks constitutes the first two stages of our progressive learning. In the final stage, we fine-tune the models using ground truth pairs. This fine-tuning process reduces the visual discrepancy between auto-generated transformations and those used to produce query videos, thereby further enhancing performance.

In summary, this paper makes the following contributions:
\begin{enumerate}
 \item We introduce a feature-compatible learning for VCD, enabling ensembles at the feature level.
 \item We effectively employ ground truth pairs to fine-tune the models, which, in conjunction with feature-compatible learning, forms progressive learning.
 \item The high-ranking results demonstrate the efficacy of our proposed FCPL method.
\end{enumerate}
\begin{figure}[t]
	\centering
	\includegraphics[width=8cm]{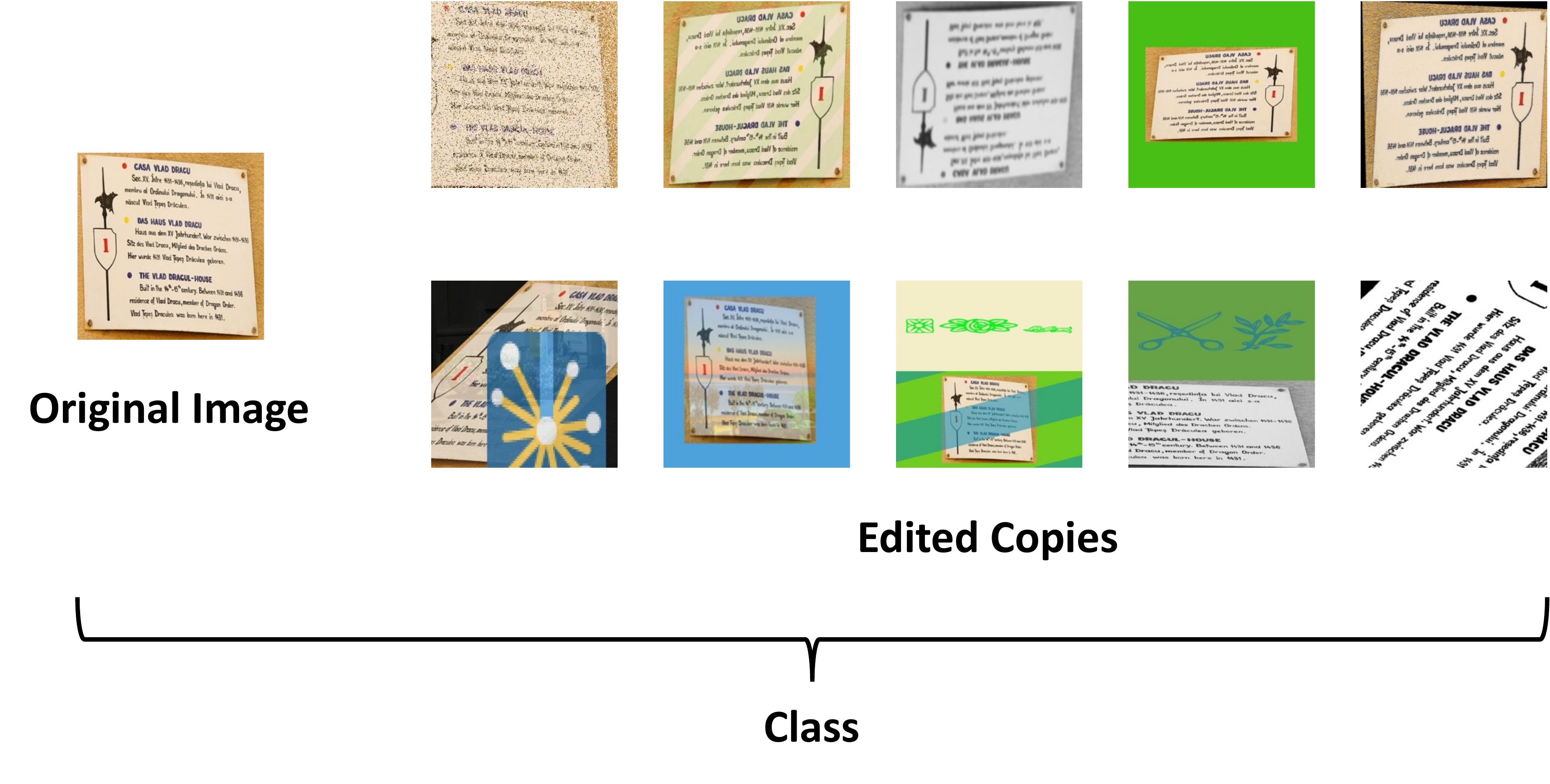}
	\caption{The demonstration for generating edited copies. An original image and its edited copies form a training class.}
	\vspace{-4pt}
	\label{Fig: S1}
\end{figure}

\section{Proposed Method}
Our FCPL comprises three stages: initial base training (utilizing ISC21 training data), feature-compatible learning (employing ISC21 training data), and fine-tuning with ground truth pairs (leveraging both ISC21 and VSC22 training data). All the training stages are on the image level rather than the video level.
\subsection{Initial Base Training}
\textbf{Generate edited copies.} Given the original image, we use pre-defined transformations to generate a training dataset. Specifically, we randomly select various transformations and utilize them to convert the original image from ISC21 \cite{douze20212021} into multiple modified versions. The original image and its edited copies together comprise a training class. A demonstration is shown in Fig. \ref{Fig: S1}. \par

\textbf{Perform deep metric learning.} Utilizing auto-generated training classes, we perform deep metric learning to train base network. This can be achieved using pairwise training \cite{hermans2017defense,sohn2016improved}, classification training \cite{liu2016large,sun2020circle,wang2018cosface}, or a combination of both methods. To simplify the process, we exclusively use CosFace \cite{wang2018cosface} as our loss function, denoted as $\mathcal{L}_{mtr}$, to train the base network.
\subsection{Feature-compatible Learning}
\begin{figure}[t]
	\centering
	\includegraphics[width=7cm]{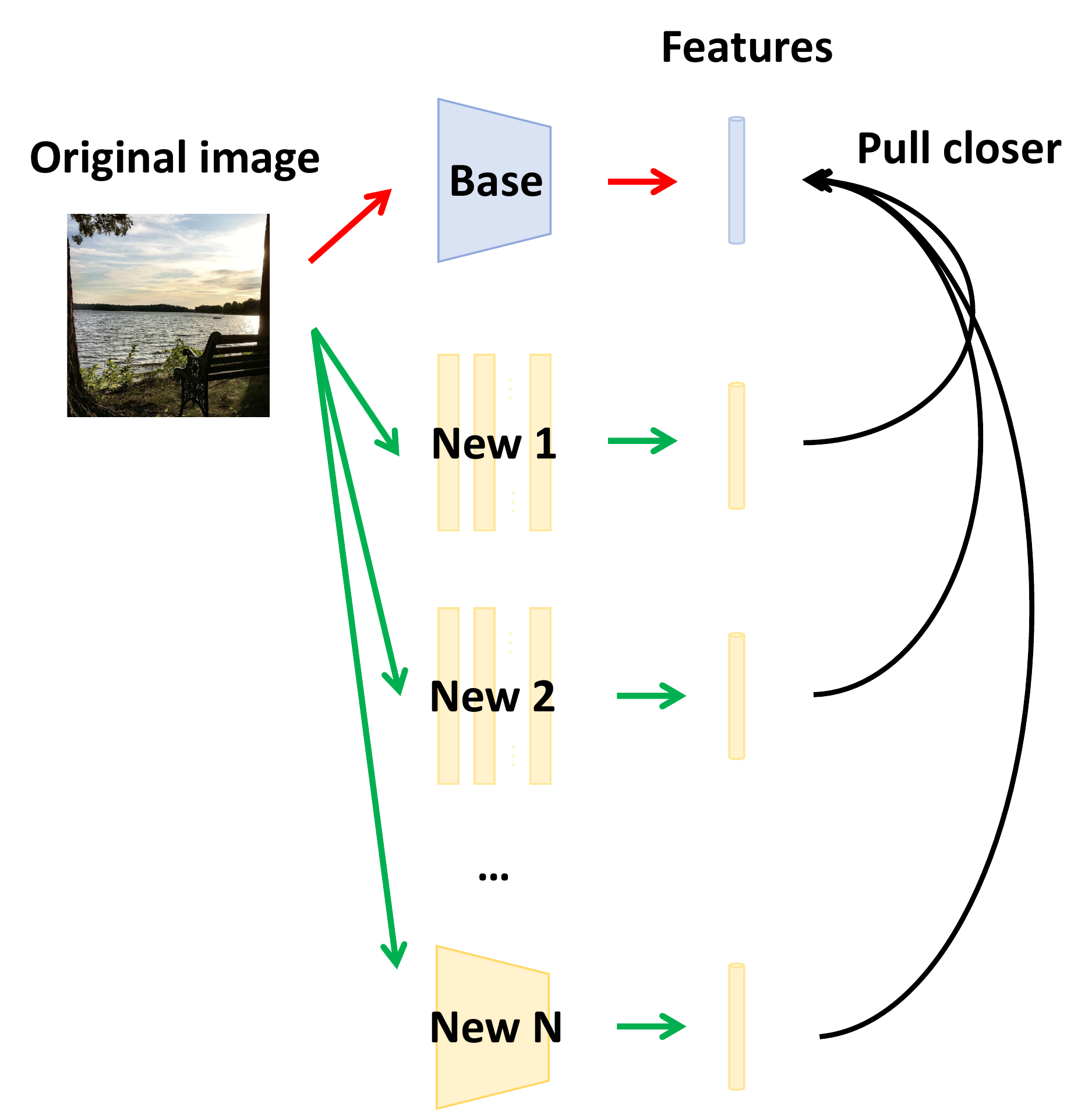}
	\caption{The demonstration of the proposed feature-compatible learning. During the training process, features extracted from original images by the new network are pulled closer to those extracted by the base network.}
	\label{Fig: S2}
\end{figure}

As depicted in Fig. \ref{Fig: S2}, we introduce a training approach called feature-compatible learning. \par

\begin{figure}[t]
	\centering
	\includegraphics[width=8cm]{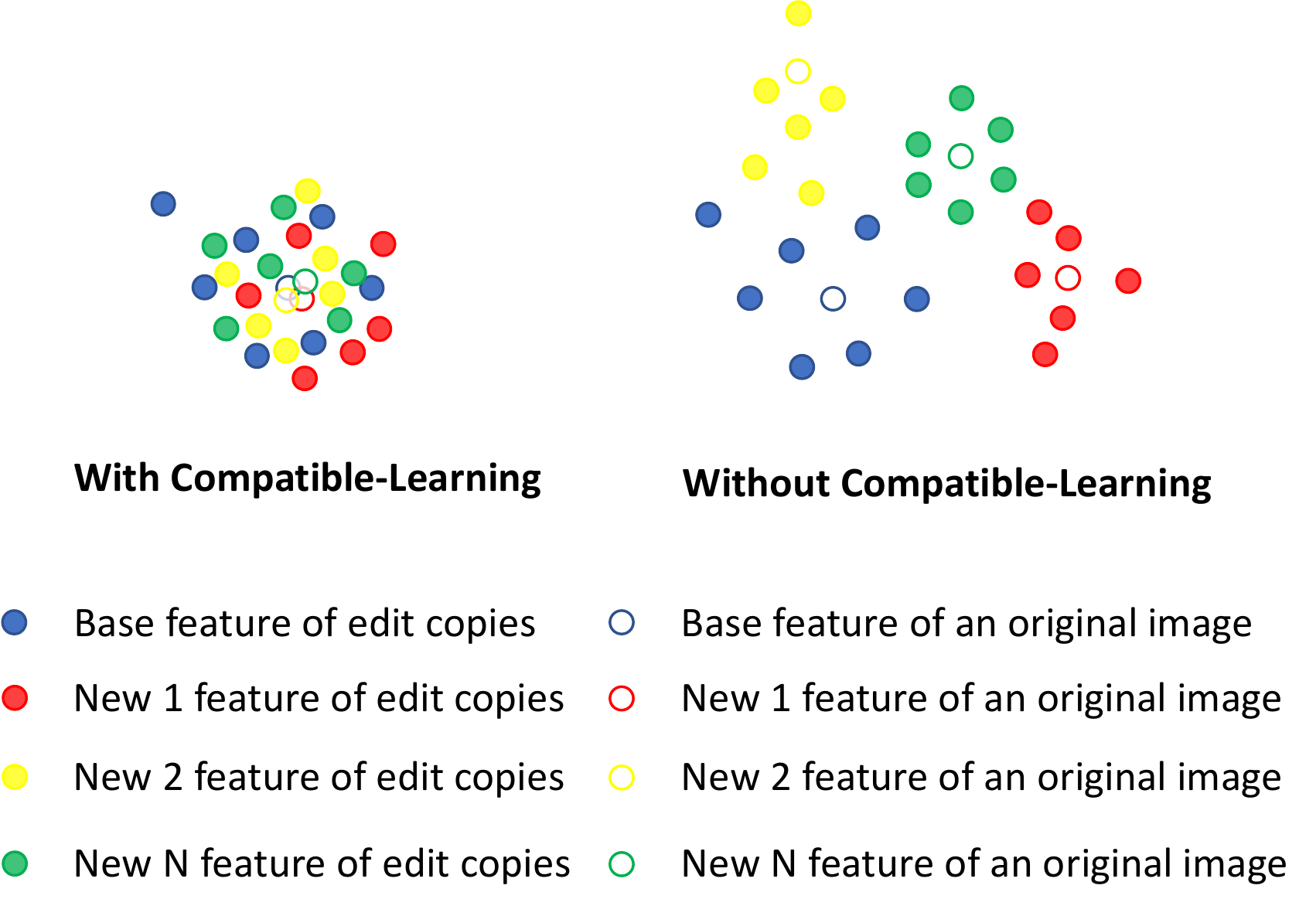}
	\caption{The comparison between with and without feature-compatible learning. With feature-compatible learning, the features gained by different models are compatible.}
	\label{Fig: reg}
\end{figure}
\begin{figure}[t]
	\centering
	\includegraphics[width=7.5cm]{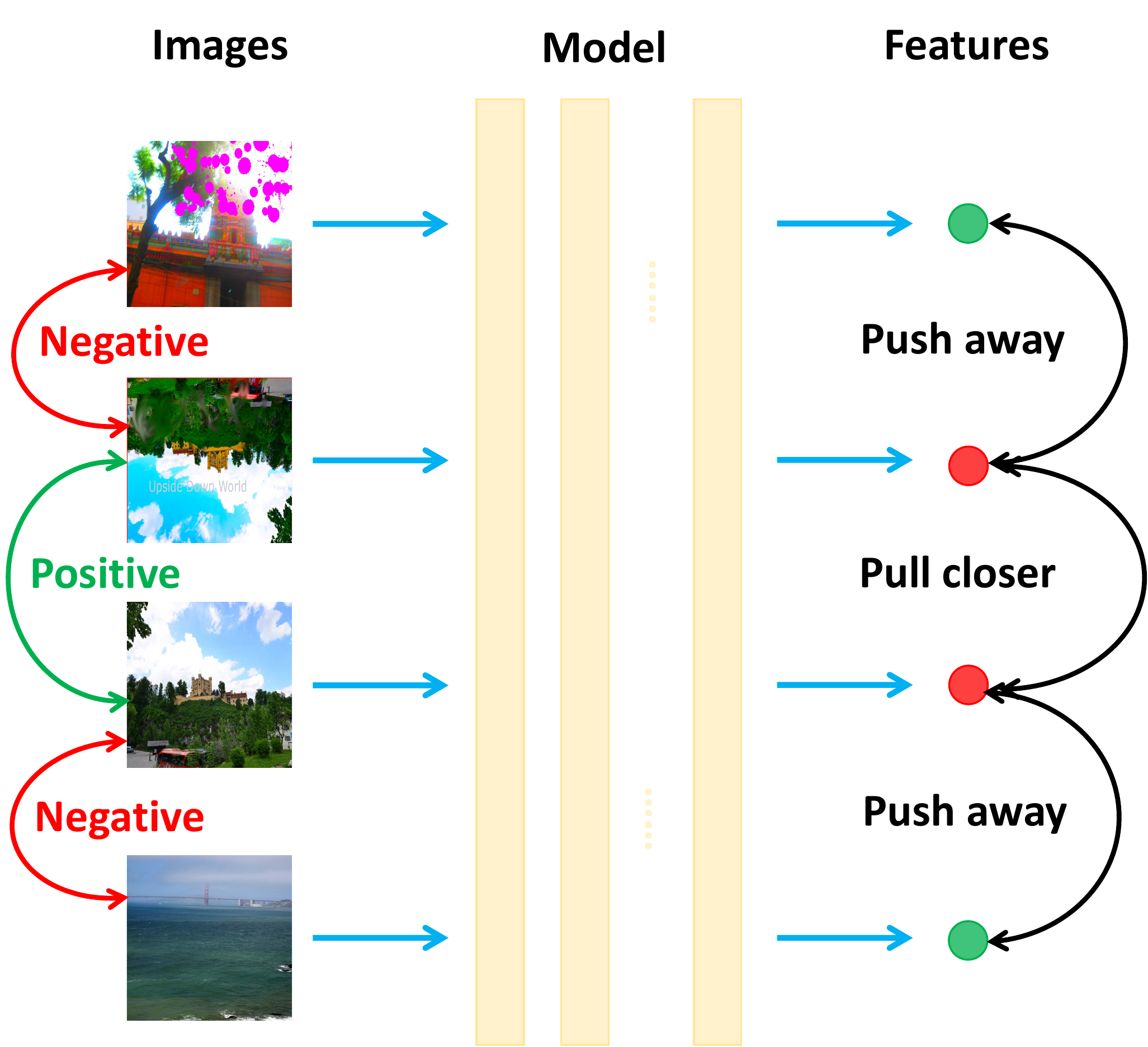}
	\caption{The demonstration of fine-tuning with the ground truth pairs.}
	\label{Fig: S3}
\end{figure}
\begin{figure*}[t]
	\centering
	\includegraphics[width=15cm]{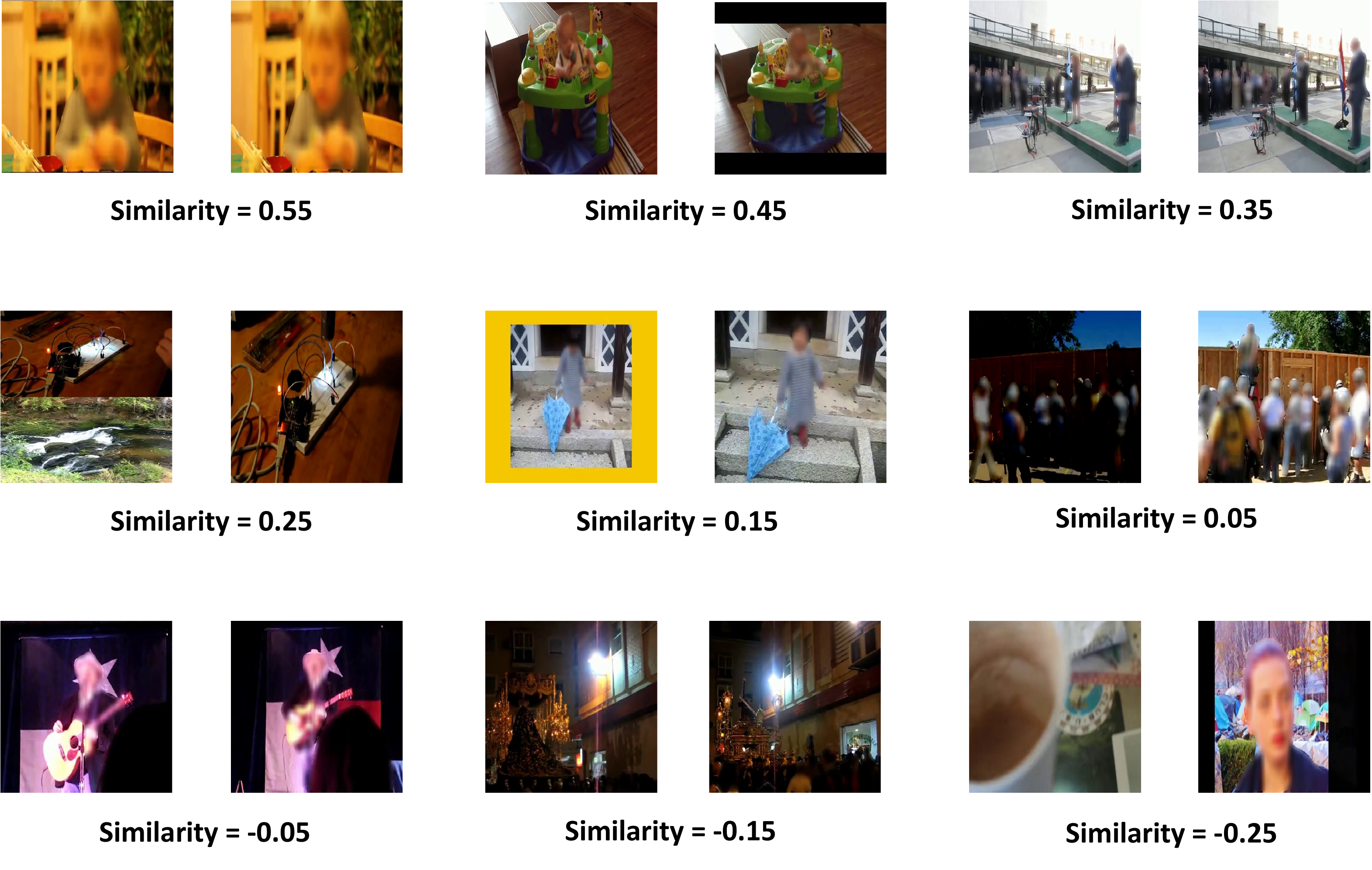}
	\caption{The visualization of matching results. In each pair, the query frame is on the left, while the reference frame is on the right.}
	\label{Fig: Vis}
\end{figure*}

Denote the original image as $x_o$, the base network as $f$, and a new network as $g$. The features of the original image extracted by the base and new network can be represented by $f\left(x_o\right)$ and $g\left(x_o\right)$, respectively. We use the $L_2$ loss to achieve compatibility:

\begin{equation}
\mathcal{L}_{com} \  =\  \sum\nolimits^{N}_{i=0} \parallel \frac{f\left( x_{o_{i}}\right)  }{\| f\left( x_{o_{i}}\right)  \|_{2} } -\frac{g\left( x_{o_{i}}\right)  }{\| g\left( x_{o_{i}}\right)  \|_{2} } \parallel_{2},
\end{equation}
where: $N$ is the number of the original images, and $\|\cdot\|_2$ is $L_2$ normalization. Therefore, when performing feature-compatible learning, the final loss is:
\begin{equation}
\mathcal{L}_{final} = \mathcal{L}_{mtr} + \lambda_r \cdot \mathcal{L}_{com},
\end{equation}
where $\lambda_r$ is the balance parameter. \par
With this learning method, we can train $N$ different backbones respectively. In practice, we choose ResNet-50 \cite{he2016deep}, ResNeXt-50 \cite{xie2017aggregated}, SKNet-50 \cite{li2019selective}, ViT \cite{dosovitskiy2020vit}, Swin Transformer \cite{liu2021Swin}, and T2T-ViT \cite{Yuan_2021_ICCV} as the new networks; and CotNet-50 \cite{cotnet} for the base one. They are both initialized by ImageNet-pre-trained models. A comparison of using feature-compatible learning versus not using it can be seen in Fig. \ref{Fig: reg}. During testing, the feature of query image $q$ is represented by $\frac{1}{N} \sum^{N}_{i=1} g_{i}\left( q\right)$, and the the feature of reference image $r$ is represented by $\frac{1}{N} \sum^{N}_{i=1} g_{i}\left( r\right)$. The feature-compatible learning ensures the ensemble at the feature level.

\subsection{Fine-tuning with Ground Truth Pairs}

Our findings indicate that the auto-generated transformations exhibit visual discrepancies with query images in the test set. As a result, we aim to employ the labeled ground truth pairs, as shown in Fig. \ref{Fig: S3}.\par
Denote a trained network as $g_t$, two images in a positive pairs as $x^1_p$ and $x^2_p$, and $x^j_n$ as the hardest negative of $x^j_p$ ($j = 1, 2$). Therefore, we have the training objectives:
\begin{equation}
\mathcal{L}_{pos} \  =\  \sum\nolimits^{M}_{i=0} \parallel \frac{g_{t}\left( x^{1}_{{}p_{i}}\right)  }{\| g_{t}\left( x^{1}_{p_{i}}\right)  \|_{2} } -\frac{g_{t}\left( x^{2}_{p_{i}}\right)  }{\| g_{t}\left( x^{2}_{p_{i}}\right)  \|_{2} } \parallel_{2},
\end{equation}
\begin{align}
\mathcal{L}_{neg} &= \frac{1}{2} \sum\nolimits^{M}_{i=0} \Bigg(\Big\| \frac{g_{t}\left( x^{1}_{p_{i}}\right)}{\| g_{t}\left( x^{1}_{p_{i}}\right) \|_{2} } -\frac{g_{t}\left( x^{1}_{n_{i}}\right)}{\| g_{t}\left( x^{1}_{n_{i}}\right) \|_{2} } \Big\|_{2} \\
&\phantom{=} +\Big\| \frac{g_{t}\left( x^{2}_{p_{i}}\right)}{\| g_{t}\left( x^{2}_{p_{i}}\right) \|_{2} } -\frac{g_{t}\left( x^{2}_{n_{i}}\right)}{\| g_{t}\left( x^{2}_{n_{i}}\right) \|_{2} } \Big\|_{2} \Bigg),
\end{align}
\begin{equation}
\mathcal{L}_{final} \  =\  \mathcal{L}_{mtr} +\lambda_{r} \cdot \mathcal{L}_{com} +\  \lambda_{pn} \cdot \left( \mathcal{L}_{pos} -\mathcal{L}_{neg} \right), 
\end{equation}
where $M$ is the number of the positive pairs, and $\lambda_{pn}$ is the balance parameter. \par

In the competition, we convert the provided positive video pairs into even more positive image pairs based on their timestamps.

\subsection{Test}
During the testing phase, we employ the ensemble feature for the descriptor track, and utilize the official TN method \cite{tan2009scalable} for localizing copy segments in the matching track.
\section{Experiments}

\subsection{Visualization}
In Fig. \ref{Fig: Vis}, we visualize some matching image pairs in the descriptor track. When the similarity score is greater than 0, the matching results appear reasonable. Interestingly, from the perspective of image copy detection, the matching pairs might not be considered true matches, as the two images could be the same instance captured at different times. Nonetheless, for the descriptor track in VCD, this distinction is not important; our primary concern is whether the two videos can be matched.

\subsection{Comparison with State-of-the-Arts}
In Tables \ref{Table: SOTA_des} and \ref{Table: SOTA_mat}, we compare the results of our method with those of other competitors in Phase 2 for both tracks. In the descriptor track, our FCPL demonstrates a performance gap of about $-2\%$ compared to the first place competitor. However, it is intriguing to note that in the matching track, we lag by more than $14\%$. We suspect this is because we only employ the traditional localization method (TN \cite{tan2009scalable}), which does not yield optimal results. Combining our FCPL with the top team's localization approach might lead to better performance.

\begin{table}[t]
\caption{The comparison between our method and others in the descriptor track.}  
\vspace*{2mm}
\small
  \begin{tabularx}{\hsize}{Y|Y|}
    \hline
Team &$\mu AP$ ($\%$) $\uparrow$\\ \hline
do something&$87.17$\\ 
\textbf{FriendshipFirst (Ours)}&$\textbf{85.14}$\\ 
cvl-descriptor &$83.62$\\
Zihao &$77.29$\\
People-AI &$68.84$\\ 
Baseline &$60.47$\\  
...&...\\
 \hline
  \end{tabularx}
  \label{Table: SOTA_des}
  \\
\end{table}

\begin{table}[t]
\caption{The comparison between our method and others in the matching track.}  
\vspace*{2mm}
\small
  \begin{tabularx}{\hsize}{Y|Y|}
    \hline
Team &$\mu AP$ ($\%$) $\uparrow$\\ \hline
do something more & $91.53$ \\
\textbf{CompetitionSecond (Ours)}&$\textbf{77.11}$\\ 
cvl-matching&$70.36$\\ 
People--AI &$50.72$\\
Baseline &$44.11$\\  
...&...\\
 \hline
  \end{tabularx}
  \label{Table: SOTA_mat}
  \\
\end{table}

\section{Conclusion}
This report introduces the Feature-Compatible Progressive Learning (FCPL) approach for Video Copy Detection (VCD). By implementing feature-compatible learning, we effectively achieve ensemble at the feature level. Progressive learning and fine-tuning on the ground truth pairs allow us to gradually enhance performance. Utilizing these techniques, we achieve the second place in both tracks of VSC22. It is unfortunate that our performance lags significantly in the matching track competition, which may be partially attributed to our reliance on traditional localizing methods.

\bibliographystyle{plain}
\bibliography{wenhao_bib}
\end{document}